\documentclass[10pt,journal,compsoc]{IEEEtran}
\ifCLASSOPTIONcompsoc
  \usepackage[nocompress]{cite}
\else
  \usepackage{cite}
\fi

\usepackage{graphicx }
\usepackage{subfigure}
\usepackage{bm}
\usepackage{amsmath, mathtools}
\usepackage{amssymb}
\usepackage{balance}
\usepackage{algorithm}
\usepackage{algorithmic}
\usepackage{colortbl}
\usepackage{multirow,tabularx}
\usepackage{booktabs}
\usepackage{caption}
\usepackage{framed}
\usepackage{balance}

\ifCLASSINFOpdf
\else
\fi

\begin{document}
%

\title{Gradient Amplification: An efficient way to train deep neural networks} 
%
%
%
%

\author{Sunitha~Basodi, 
        Chunyan~Ji, 
Haiping~Zhang,
and Yi~Pan\IEEEauthorrefmark{1}



\IEEEcompsocitemizethanks{\IEEEcompsocthanksitem Sunitha~Basodi, Chunyan~Ji and Yi~Pan are with the Department of Computer Science, Georgia State University, Atlanta, GA 30302, USA. E-mail: sbasodi1@student.gsu.edu;cji2@student.gsu.edu; yipan@gsu.edu. \protect\\ 
\IEEEcompsocthanksitem Haiping~Zhang is with the Center for High Performance Computing, Joint Engineering Research Center for Health Big Data Intelligent Analysis Technology, Shenzhen Institutes of Advanced Technology Chinese Academy of Sciences, Shenzhen, Guangdong, PR China 518 055. 

\IEEEcompsocthanksitem *
To whom correspondence should be addressed (yipan@gsu.edu).}
}
\IEEEtitleabstractindextext{%
\begin{abstract}

Improving performance of deep learning models and reducing their training times are ongoing challenges in deep neural networks. There are several approaches proposed to address these challenges one of which is to increase the depth of the neural networks. Such deeper networks not only increase training times, but also suffer from vanishing gradients problem while training. In this work, we propose gradient amplification approach for training deep learning models to prevent vanishing gradients and also develop a training strategy to enable or disable gradient amplification method across several epochs with different learning rates. We perform experiments on VGG-19 and resnet (Resnet-18 and Resnet-34) models, 
and study the impact of amplification parameters on these models in detail. Our proposed approach improves performance of these deep learning models even at higher learning rates, thereby allowing these models to achieve higher performance with reduced training time. 
\end{abstract}

}

\thispagestyle{empty}
\clearpage
\pagenumbering{arabic} 
\maketitle

\IEEEdisplaynontitleabstractindextext

%
\IEEEpeerreviewmaketitle

\IEEEraisesectionheading{\section{Introduction}\label{sec:introduction}}

%
%
%
%
Deep learning models have achieved state-of-the-art performances in several areas including computer vision\cite{he2016deep}, automatic speech recognition\cite{hinton2012deep}, natural language processing\cite{hochreiter1997long} and beyond\cite{kamilaris2018deep,litjens2017survey, zhang2018survey, zhang2018deep, wang2019deep}. These models are designed, trained, and tuned to achieve better performance for a given dataset. Their performance increases further with the increase in the depth of the network\cite{huang2016deep}. The major challenges associated with the increase in the network architecture is the high amount of time required to train the model even on parallel computation resources and also vanishing gradients\cite{huang2016deep}. 
Training deep neural networks is time-consuming, which could take days or sometimes weeks depending on the type of the model architecture and size of the dataset. One way to speed up the training process is to increase the learning rate. This will accelerate the training process by quickly converging to optima, but also has the risk of missing the global optima resulting in sub-optimal solutions or sometimes non-convergence\cite{art2_learningrate_vs_time}. Lower learning rates does not have such a risk and can converge to optima, but increases training speeds. In general, training process with a learning rate scheduler begins with higher learning rates for a few epochs, followed by reduction of learning rates for the next couple of epochs; which is repeated until the desired optima or model performance is achieved. One way to improve the training speed of deep learning models can be to determine ways to achieve optimal model parameters at larger learning rates.

The other important area of research in deep learning models is to prevent vanishing gradient problem\cite{hochreiter2001gradient, goh2017deep, hanin2018neural}. The vanishing gradient problem occurs during training of artificial neural networks, specifically during backpropagation. There are several approaches to avoid this problem. One suggested early method was to perform a two step training process which involves network weight initialization followed by fine-tuning using backpropagation method \cite{schmidhuber1992learning}. The other simpler methods that prevent this problem are Rectified Linear Unit (ReLU) activation function \cite{nair2010rectified, glorot2011deep} and batch normalization (BN) \cite{ioffe2015batch}. Since ReLU activation saturates inputs in only one direction, therefore has less impact of vanishing gradients. The other recent approach of batch normalization not only improves the performance of the model, but also reduces vanishing gradient problems. Resnet architecture have residual connections which also overcome vanishing gradient problem to some extent\cite{he2016deep}. Lately, due to the improvement of hardware along with the computational abilities of Graphical Processing Units (GPUs), neural networks can be trained without the issue of vanishing gradients. 

In this work, we propose a novel gradient amplification approach along with a training strategy which addresses the challenges discussed above. In this method, gradients are dynamically increased for some layers during backpropagation so that significant gradient values are propagated to the initial layers. This process is repeated for a few epochs along with the normal training process with no gradient amplification for the other epochs. When neural networks are trained using this method, we observe that the testing/training accuracies of the models improve and achieve higher accuracies faster, even at higher learning rates, and therefore reduces the training time of these deep learning models. 


Our contributions include the following: 
\begin{itemize}
\renewcommand\labelitemi{--}
\item We propose a novel way of amplifying gradients during backpropagation for effective training of deep neural networks 
\item We suggest a unique training strategy which includes amplification during certain epochs along with normal training with no amplification
\item We perform comprehensive experiments to understand the impact of different parameters used in amplification
\item We perform step-wise analysis of training strategy demonstrating the best strategy with different learning rates
\end{itemize}


The remainder of this paper is organized as follows. Related works are briefly described in Section \ref{sec:relatedwork}. Our proposed approach is presented in Section \ref{sec:methods}. Experimental setup, results and their comparisons are covered in Section \ref{sec:experiments}, followed by conclusions in Section \ref{sec:conclusion}.

\section{Background } \label{sec:relatedwork}
In this section, we briefly discuss the existing approaches to address vanishing gradient problem, reduce the training time of deep learning models, and the impact of learning rates. 

\subsection{Vanishing gradients}
Vanishing gradient problem\cite{hochreiter2001gradient, goh2017deep, hanin2018neural} occurs while training artificial neural networks during backpropagation and can become significant with the increase of depth of the network. In gradient-based learning methods, during backpropagation, network weights are updated proportional to the gradient value (partial derivative of the cost function with respect to the current weights) after each training iteration (epoch). Depending on the type of the activation functions and network architectures, sometimes the gradient value is too small and its value gets gradually diminished during backpropagation to the initial layers. This prevents the network from updating its weights and also sometimes when the value is too small, the network may be completely stopped from training (updating weights). Though there is no fundamental solution to this problem, but some of the approaches help to avoid it\cite{schmidhuber2015deep}. One such approach consists of performing a two step training process. In the first step, network weights are trained using unsupervised learning methods (such as auto-encoding) and then the weights are fine-tuned using backpropagation method\cite{schmidhuber1992learning}. Other simpler methods that prevent this problem are ReLU activation function \cite{nair2010rectified, glorot2011deep}, batch normalization(BN)\cite{ioffe2015batch} and Resnet networks\cite{he2016deep}. ReLU activation zeros the negative values and only considers positive values. As it saturates inputs in only one direction, it has less impact of vanishing gradients. The other approach, batch normalization, also reduces vanishing gradient problems other than boosting the performance of the model. In batch normalization, during every training iteration, the input data is normalized to reduce its variance, so that the data does not have large bounds. Since inputs are normalized, gradients are also regulated\cite{ioffe2015batch}. Resnet network architectures have residual networks have residual connections which help to improve on this problem. In addition to these approaches, recent advancement in the hardware has also played a crucial role in solving this issue. Increased computational abilities and availability of GPUs aid in reducing this problem. 

\subsection{Learning rates}
Learning rate is one of the most important hyperparameters which controls the performance of deep neural networks. Having higher learning rates cause the model to train faster but might have sub-optimal solutions. However, lower learning rates take longer time to train the model, but can achieve better optimal solutions\cite{art2_learningrate_vs_time}. There are several approaches designed to take advantage of them. One such method is learning rate scheduler where we start with higher learning rates and gradually lower the rates with training epochs \cite{darken1992learning}. There are several ways in which such a scheduler can be designed, namely, directly assigning the learning rates to the epochs, gradually decaying the learning rate based on the current learning rate, current epoch and total number of epochs(time-based decay); reducing the learning rate in a step-wise manner after a certain number of epochs(step decay); and exponentially decaying the learning rate based on the initial learning rate and the current epoch(exponential decay). Another approach include adapting learning rate dynamically based on the performance of the optimization algorithm without need of any scheduling, some of such methods include Adagrad\cite{duchi2011adaptive},  Adadelta\cite{zeiler2012adadelta}, RMSprop\cite{graves2013generating} and Adam \cite{kingma2014adam}. Article \cite{art1_learningrate} summarizes all the above discussed methods in detail. Paper \cite{schaul2013no} proposes a method to automatic tune the learning rate based on the local gradient variations of the data samples which has similar performance to other adaptive learning rate methods. Another paper \cite{smith2017don} shows that models can achieve similar test performance with out decaying the learning rate but by increasing the batch size instead. This method not only has fewer parameter updates but also increases parallelism thereby reducing training times.  


In the next section, we discuss our proposed method and also training strategy with a fixed learning rate schedule across epochs which achieves better accuracies even at the higher values of learning rates.

\section{ Proposed Method }\label{sec:methods}

Our proposed approach is to dynamically amplify (increase) the value of the gradients for a selection of layers during backpropagation. This ensures that the gradient values are not diminished while updating weights for the initial network layers and a significant value of the gradients is available during backpropagation even for deep neural networks with large number of layers. 
Architectures of neural networks have evolved over the years and there are many different layers where such an amplification can be done. The layers on which gradient amplification can be performed during backpropagation are arranged into a group, say $G$. To determine this group, firstly, the type of layers that needs to be included for gradient amplification should be identified. Each of the layers such as convolution layers, batch normalization layers, pooling layers, activation function layers and so on can be chosen to be included in the group. The type of the layer considered plays a crucial role in the performance of the model. Gradient amplification is done on a subset of the layers from this group $G$, which we refer as $amp$ layers in the rest of the paper. Selection of the $amp$ layers from a group of layers can be done in various methods. In this work, we determine the $amp$ layers by random selection. To identify which subset size has better performance, we choose a parameter $\beta$ representing the ratio of $amp$ layers to be selected from all the layers in the group $G$. Gradients are amplified when they pass through these randomly selected layers during backpropagation. During amplification, value of gradients is increased at run time by multiplying the actual gradient values by a factor $\Gamma$. The value of $\Gamma$ is important as it should not be too small or too large. If the value of $\Gamma$ is too small, then the increase might not be effective and if it is too large it might overfit the data or cause incorrect weight updates. During training, we perform gradient amplification for some epochs and with no gradient amplification for other epochs. Algorithm \ref{alg:amp_training} describes the training process with gradient amplification and algorithm \ref{alg:find_amp} describes the steps for the selection of layers from $G$.

\begin{algorithm}
 \caption{Training process with gradient amplification}
 \label{alg:amp_training}
 \begin{algorithmic}
 \renewcommand{\algorithmicrequire}{\textbf{Input:}}
 \REQUIRE $M$, $params$=[($e_{1}, \eta_{1}, \beta_{1}, \Gamma_{1}$), ($e_{2}, \eta_{2}, \beta_{2}, \Gamma_{2}$), $...$]
\\
 \emph{Variables:}\\

 $\Gamma$ is gradient amplification factor \\
 $\beta$ is ratio of layers to be selected for amplification\\
 $\eta$ is the learning rate \\
 $amp$ the set of layers selected to perform amplification \\
 $M$ is the neural network model\\

 $params$ is an array of elements, each in the format ($end\_epoch$, $\eta$, $\beta$, $\Gamma$) 

 \emph{ }\\
 
 $start\_epoch$=0
   \FOR {$(e_i, \eta_i , \beta_i, \Gamma_i)$ in  $params$}
   \STATE    \textit{update learning rate to $\eta_i$  }
   \STATE optimizer=sdg\_optimizer($\eta_i$)
   \IF {($\beta_i > 0$)}
   \STATE $amp$ =  \textsc{GetGradientAmpLayers({$M$, $\beta$ })}
  \ENDIF

   \FOR {$k = start\_epoch $ to $e_i$} 
   \STATE train the model $M$
   \IF {($\beta_i > 0$)}
   \STATE multiply gradients with $\Gamma_i$ for layers in $amp$ during backpropagation
    
\ELSE 
\STATE {perform regular backpropagation without gradient amplification}

  \ENDIF

  \ENDFOR
   \STATE  $start\_epoch$=$e_i$
   \STATE reset $amp$
   \STATE evaluate model $M$ with a testing set
   \ENDFOR

 \RETURN{}
  \end{algorithmic} 
 \end{algorithm}

\begin{algorithm}
\caption{Determination of $amp$ layers}
\label{alg:find_amp}
\begin{algorithmic}
\renewcommand{\algorithmicrequire}{\textbf{Input:}}

\REQUIRE in $M$, $\beta$

 $\beta$ is ratio of layers to be selected for amplification\\
 $G$ is a set consisting of a group of all layers that can be used for gradient amplification \\
 $layer\_types$= Set indicating the type of layers to be used for amplification\\
 
  \STATE Function \textsc{ GetGradientAmpLayers({$M$, $\beta$ }) } \label{alg:a}
 \FORALL{ $layer$ in $layer\_types$ }
 \STATE include $layer$ in $G$  
 \ENDFOR
 \STATE $amp\_size$ = $\beta$*sizeof($G$)
 \STATE $amp$ = \textsc{RandomSelect({$G$, $amp\_size$})}
 
  \STATE EndFunction

\end{algorithmic}
\end{algorithm}

\section{Experiments \& Results}\label{sec:experiments}

\subsection{Setup}

\begin{figure}
\includegraphics[width=0.5\textwidth]{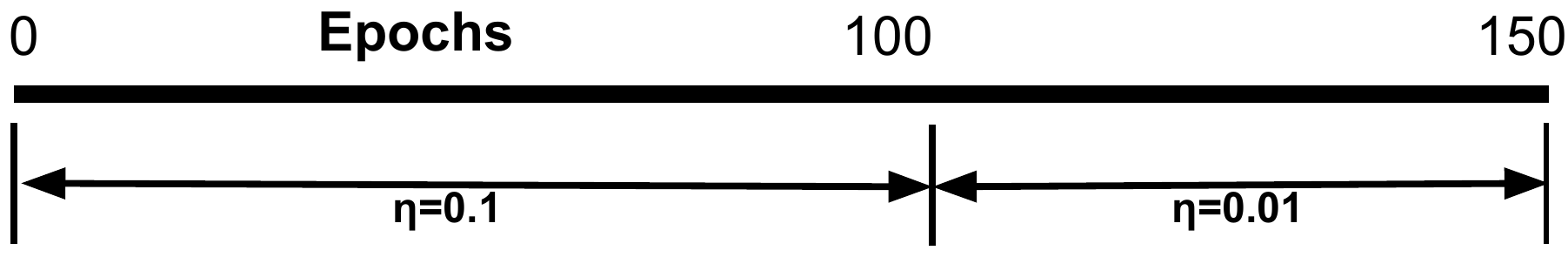}
\caption{ Experiment setting showing the number of epochs and learning rates corresponding to epochs for training all the models.}
\label{fig:GradAmp_Epochs_1}
\end{figure}

Our experiments are performed on CIFAR 10 dataset which consists of 60000 colored images of 10 classes with 6000 images per class and each image has 32x32 resolution. We implement our algorithms using python and pytorch \cite{paszke2017automatic} libraries. In our experiments, we employ several standard deep learning models and train them for 150 epochs. The number of epochs, combination of number of epochs and learning rates can be chosen as one thinks best. In this work, the first 100 epochs have learning rate of 0.1 and the next 50 epochs have the learning rate of 0.01 (as shown in Fig. \ref{fig:GradAmp_Epochs_1}. The first 50 epochs are trained with learning rate of 0.1 without gradient amplification. This is because for the first few epochs, the model is considered to be in transient phase and the network parameters undergo significant changes. This initial transient can be considered for any number of epochs and in this work, we set it to 50 epochs. The next 50 epochs have the same learning rate of 0.1 but has gradient amplification applied during backpropagation while training the model (as shown in Fig. \ref{fig:GradAmp_Epochs_2}). After identifying the best $params$ with gradient amplification for epochs $51-100$, using those $params$ for those epochs, we extend amplification for epochs $101-130$ to identify the best $params$ and with no amplification for epochs $131-150$, as shown Fig. \ref{fig:GradAmp_Epochs_3}. There are mainly three important parameters while applying gradient amplification method namely, the type of the layers to be employed for amplification,  the ratio of layers ($\beta$) to be chosen from selected layers  to perform amplification and gradient amplification factor. The effects of varying each of these parameters are explained in detail in the subsections below. We run our experiments on Resnet and VGG models with different architectures. 

Here we perform three phase analysis while evaluating our model.
	
\paragraph{Phase1} In this phase, we choose the type of layers to be considered for amplification. There are several types of layers at which amplification can be applied such as activation function layers, pooling layers, batch normalization layers and convolution layers. Convolution layers apply kernel functions and extract important features from the data and 
 pooling layers perform accumulation of features over a grid using several strategies such as retrieving maximum values, minimum values, averaging, fractional pooling and so on. Since the network parameter tuning while training can be sensitive to these values, in this work, we do not perform amplification on these layers. Batch normalization layers normalize data over a batch of inputs, and activation function layers transform data non-linearly before forwarding it to the succeeding layers. In our work, we perform gradient amplification on batch normalization and activation function layers. ReLU is the activation function used in Resnet and VGG models. From these two types of layers, either one or both of them can be considered for amplification. Once the type of the layers is selected, we now tag all the layers of the selected type to belong to the group $G$. We now move to the next phase to determine the final amplification layers $amp$.  

\paragraph{Phase2} Once the set of layers $G$ is determined, the next task is to find the subset of layers which gives better performance. It requires identifying subset size and selection of those many layers from $G$. Since the size is unknown, experiments are performed by selecting the size to be a ratio of size of $G$. This ratio, $\beta$, is chosen from the set $\beta \in \{0, 0.1, 0.2,...,0.9, 1\}$. The actual size of $amp$ is determined by the value $\beta \times size\_of(G)$. When the value is $0$, no layers are chosen and gradient amplification is not performed. When the value is $1$, then all the layers in $G$ are considered for amplification. $0$ is included to verify whether the model performs better without gradient amplification or vice versa. Random selection is employed to select $amp$ subset of layers from $G$.  We perform experiments with all these sizes and select the model with the best performance.

\paragraph{Phase3} In this phase, the layers $amp$ on which gradient amplification can be applied are known. The only parameter left to explore is $\Gamma$, the factor with which gradient needs to be amplified. To reduce computation complexity in testing all the combinations of parameter values $amp$, $\beta$ and $\Gamma$, firstly experiments are performed on all combinations of $amp$ and $\beta$ i.e., until phase-2, then the best models are chosen from phase-2 and analyzed by varying $\Gamma$. The value of $\Gamma$ is firstly varied from $\{1,2,3,..,10\}$ to analyze the impact of amplification and then fine-tuned by varying from $\{1.1,1.2,...,2.9,3.0\}$ to determine the value that works best during training.

\subsection{Results}

\newcommand{\TTCLR}[1]{\textcolor{black}{#1}}

In our experiments, we employ Resnet-18, Resnet-34 and VGG-19 models and perform thorough analysis. As the complexity of the model and the depth of the network increases, it takes longer to compute and requires more GPU/CPU resources. Since we perform many experiments (around hundreds), having models with relatively simpler architectures and less layers would make the computation time faster. Most of our experiments are performed on High Performance Computing(HPC) cluster at Georgia State University(GSU) \cite{gsu_hpc}.%

While performing experiments, we choose either batch normalization layers or ReLU layers or both and then verify their performance over multiple epochs. We first explain the training $params$ which is important to understand the performance tables. We train our models for 150 epochs and the learning rate of the first 100 epochs is 0.1 and the next 50 is 0.01. We train the models with no gradient amplification for the first 50 epochs as the initial transient and for the next epochs, we aim to identify the pattern to select the epochs which improve the overall performance of the model. We follow the training steps mentioned in Algorithm \ref{alg:amp_training} and $params$=[($e_{1}, \eta_{1}, \beta_{1}, \Gamma_{1}$), ($e_{2}, \eta_{2}, \beta_{2}, \Gamma_{2}$), $...$] is chosen as $[(50, 0.1, 0, 1), (100, 0.1, 0, 1), (130, 0.01, 0, 1), (150, 0.01, 0, \linebreak 1)]$ when no gradient amplification is performed. The values in each element represent end epoch, learning rate, ratio of amplified layers and gradient amplification factor respectively. For instance, $(50, 0.1, 0, 1)$ means that the model is trained with learning rate $0.1$ until we reach $50$ epochs, during which $0$ layers are selected for gradient amplification and amplification factor is $1$. 

\begin{figure}%
    \centering

    \subfigure[Step-1\label{fig:GradAmp_Epochs_2}]{\includegraphics[width=1\linewidth]{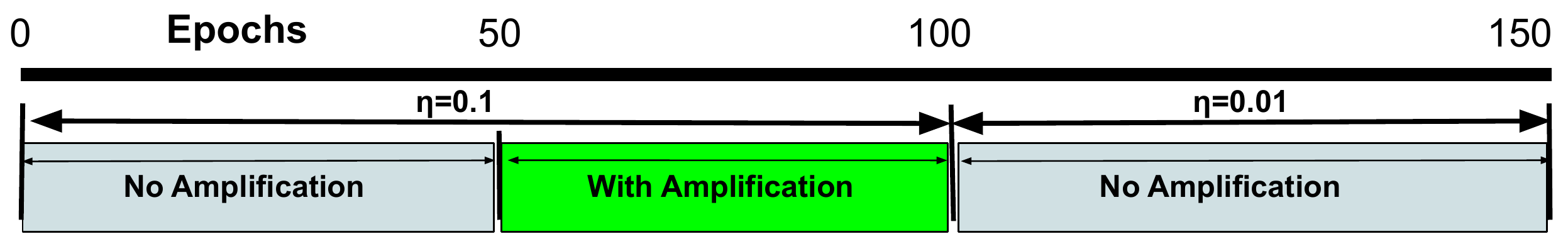}}
    \subfigure[Step-2\label{fig:GradAmp_Epochs_3}]{\includegraphics[width=1\linewidth]{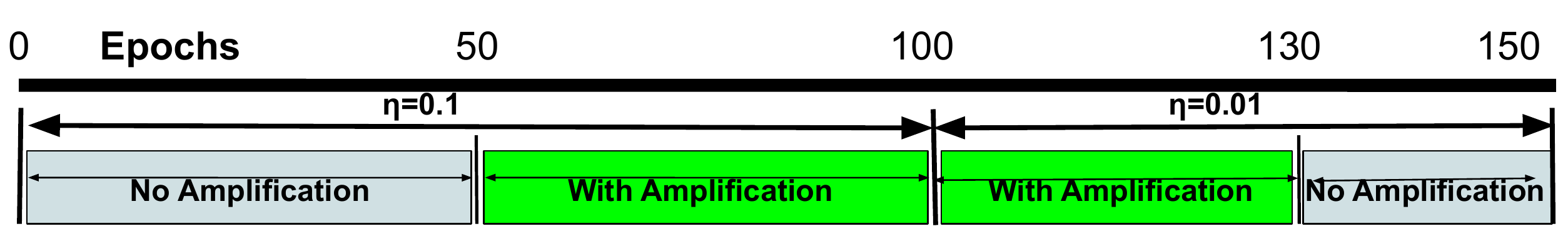}}
  
    \caption{ Two step training process carried out during performance analysis of deep learning models. Experiments are first executed on the models with training steps shown in step-1 (a). For step-2(b), ratio parameters for gradient amplification which have better performance of the models in step-1 are considered as the parameters for epochs 51-100 epochs and experiments are performed by analyzing ratio parameters for epochs 101-130, with no amplification from epochs 131-150. These settings show the number of epochs and the learning rates corresponding to these epochs while training these models.}%
    \label{fig:gradamp_epochs}%
\end{figure}

\begin{figure}
\includegraphics[width=0.5\textwidth]{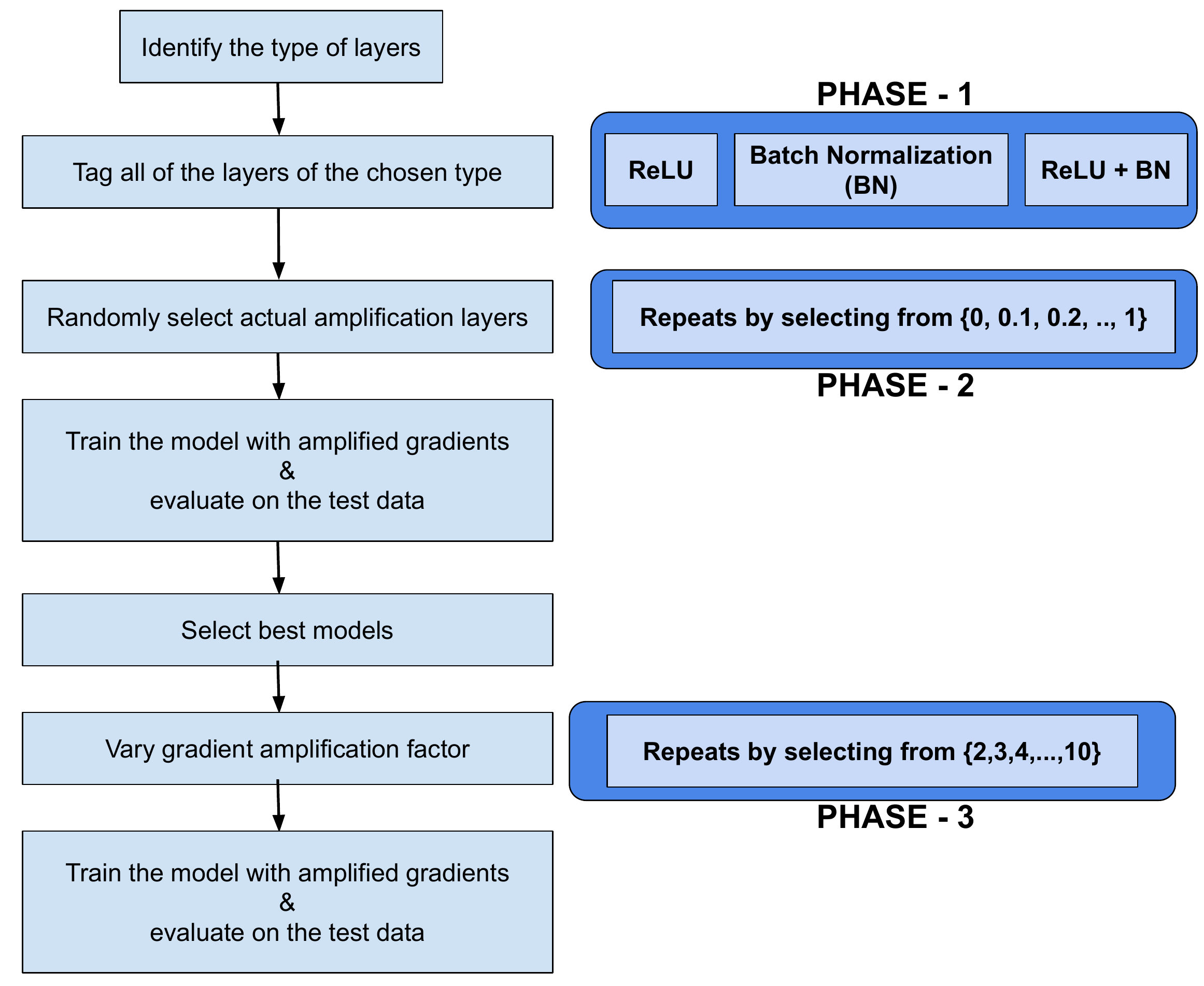}
\caption{ Overview of all the experiments performed by varying different parameters of gradient amplification.}
\label{fig:experiment_overview}
\end{figure}

Performance of original models with no gradient amplification is firstly recorded. Next, models with gradient amplification are experimented in two steps. We first set $params$ as \textbf{$[(50, 0.1, 0, 1), (100, 0.1, $\TTCLR{$xx$}$, 2), (130, 0.01, 0, 1), (150, 0.01, 0,\linebreak 1)]$}. That is, no gradient amplification is applied for the first and the last 50 epochs, as shown in Fig. \ref{fig:GradAmp_Epochs_2}. For epochs $51-100$, the ratio of selected layers is scanned from $\{0, 0.1, 0.2,...,1\}$ to identify the best model with the amplification factor $2$. For simplicity, we define $S1\_\{mm\}$ to represent the modified ratio $mm$ during epochs 51-100 in step-1 while performing amplification, and $S2\_\{mm\}\_\{nn\}$ to represent the modified ratio $mm$ during epochs 51-100  and  $nn$ during epochs 101-130, respectively, during amplification in step-2. So, the $params$ defined above will be represented as $S1\_$\TTCLR{$xx$}, where \TTCLR{$xx$} represents the value that is varied. Once we identify the best ratio for $51-100$ epochs, say $0.7$, we then run the experiments with different ratio values for the next 30 epochs by setting $params$  to be $S2\_0.7\_$\TTCLR{$xx$} i.e., $[(50, 0.1, 0, \linebreak 1), (100, 0.1, 0.7, 2), (130, 0.01,$\TTCLR{$xx$}$, 2), (150, 0.01, 0, 1)]$ as shown in Fig. \ref{fig:GradAmp_Epochs_3}. Note that, the learning rate is decreased to 0.01 after 100 epochs. After these experiments, the best models are chosen to perform experiments in phase-3 to analyze the impact of gradient amplification factor on its performance. All the phases and various experiments performed are shown in Fig. \ref{fig:experiment_overview}.  

From our initial experiments, we observe that the ratio values $\{0.1, 0.3, 0.5, 0.6\}$ on average provide better results in step-1, explained below in detail in Phase-2.  Instead of running step-2 only on the best models from step-1, different models are built with ratio values $\{0.1, 0.3, 0.5, 0.6\}$ for epochs 51-100 where the learning rate is 0.1. We perform our analysis on phase-1 and phase-2 for the following amplification $params$ in step-2 (see \ref{fig:GradAmp_Epochs_3}) : 
\begin{itemize}
  \item $S2\_0.1\_$\TTCLR{$xx$}: {\footnotesize $[(100, 0.1, \emph{0.1}, 2), (130, 0.01,$\TTCLR{$xx$}$, 2)] $ }
  \item $S2\_0.3\_$\TTCLR{$xx$}: {\footnotesize $[(100, 0.1, \emph{0.3}, 2), (130,0.01,$\TTCLR{$xx$}$, 2)]$}
  \item $S2\_0.5\_$\TTCLR{$xx$}: {\footnotesize $[(100, 0.1, \emph{0.5}, 2), (130, 0.01,$\TTCLR{$xx$}$, 2)]$}
  \item $S2\_0.6\_$\TTCLR{$xx$}: {\footnotesize $[(100, 0.1, \emph{0.6}, 2), (130, 0.01,$\TTCLR{$xx$}$, 2)]$}
\end{itemize}

\begin{figure*}
    \begin{framed}
  \centering
    \subfigure[$Step2$ for $VGG-19$\label{fig:VGG-19_step2}]{\includegraphics[width=0.33\textwidth]{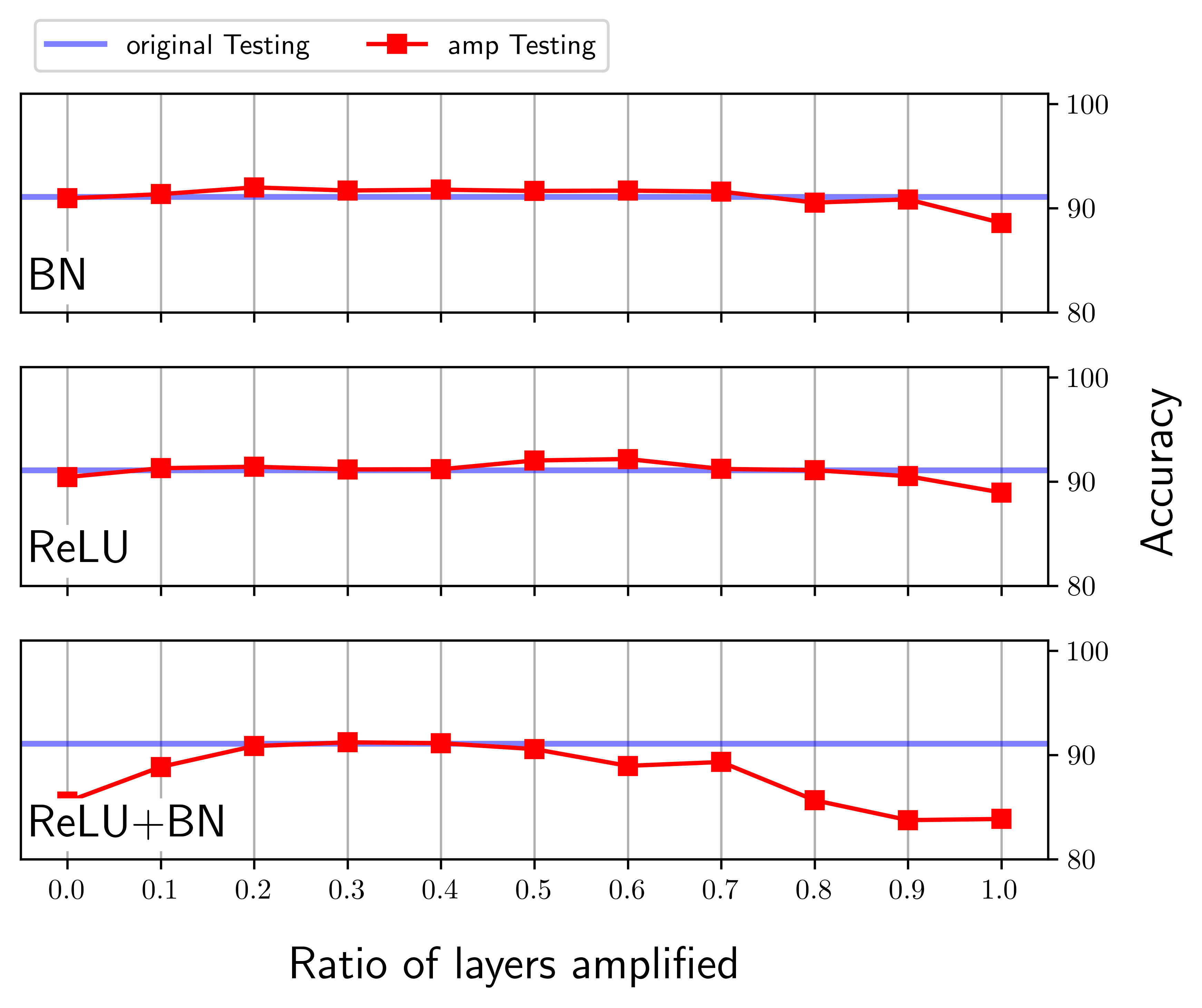}}
    \subfigure[$Step2$ for $Resnet-18$\label{fig:Resnet-18_step2}]{\includegraphics[width=0.33\textwidth]{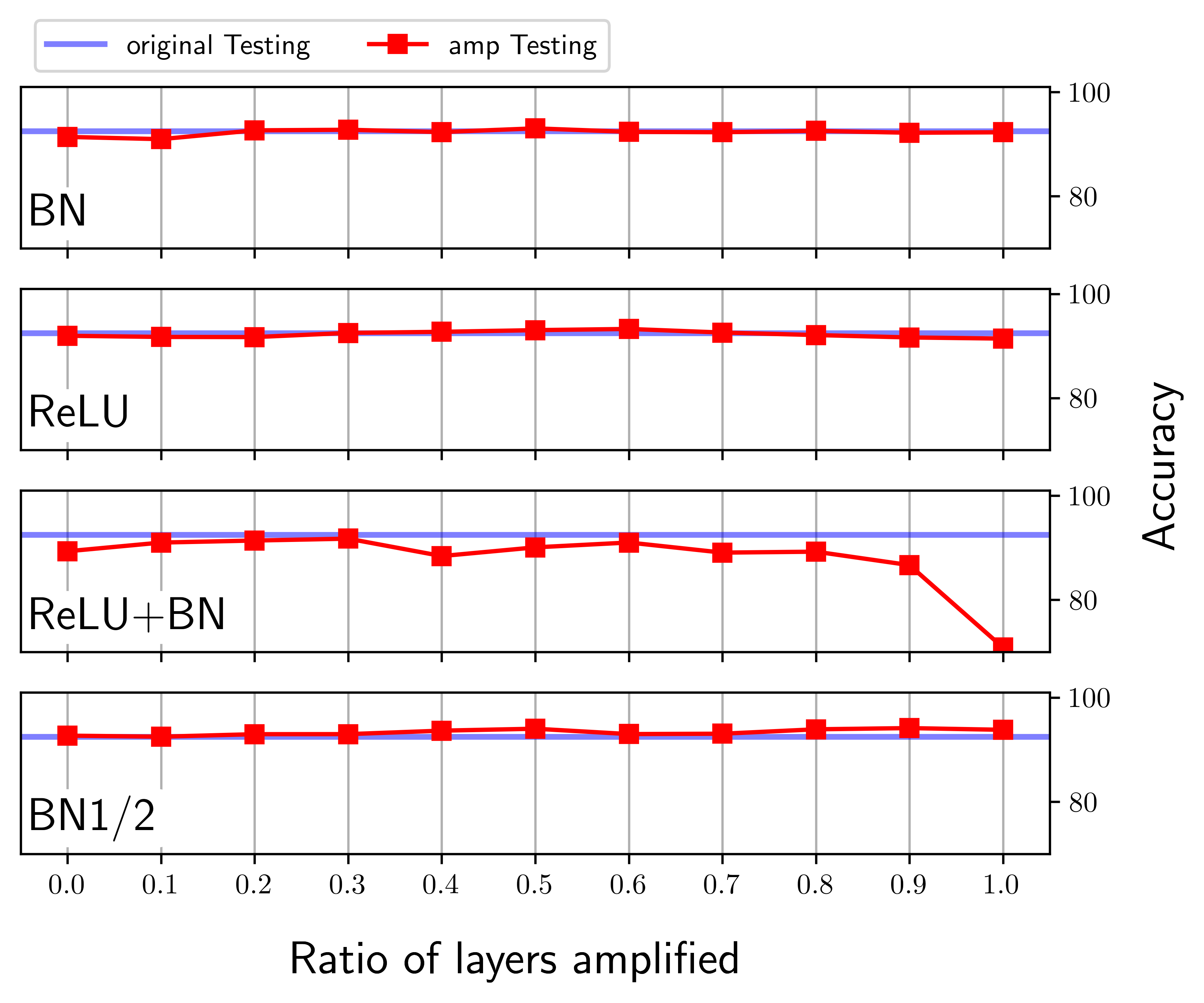}}
    \subfigure[$Step2$ for $Resnet-34$\label{fig:Resnet-34_step2}]{\includegraphics[width=0.33\textwidth]{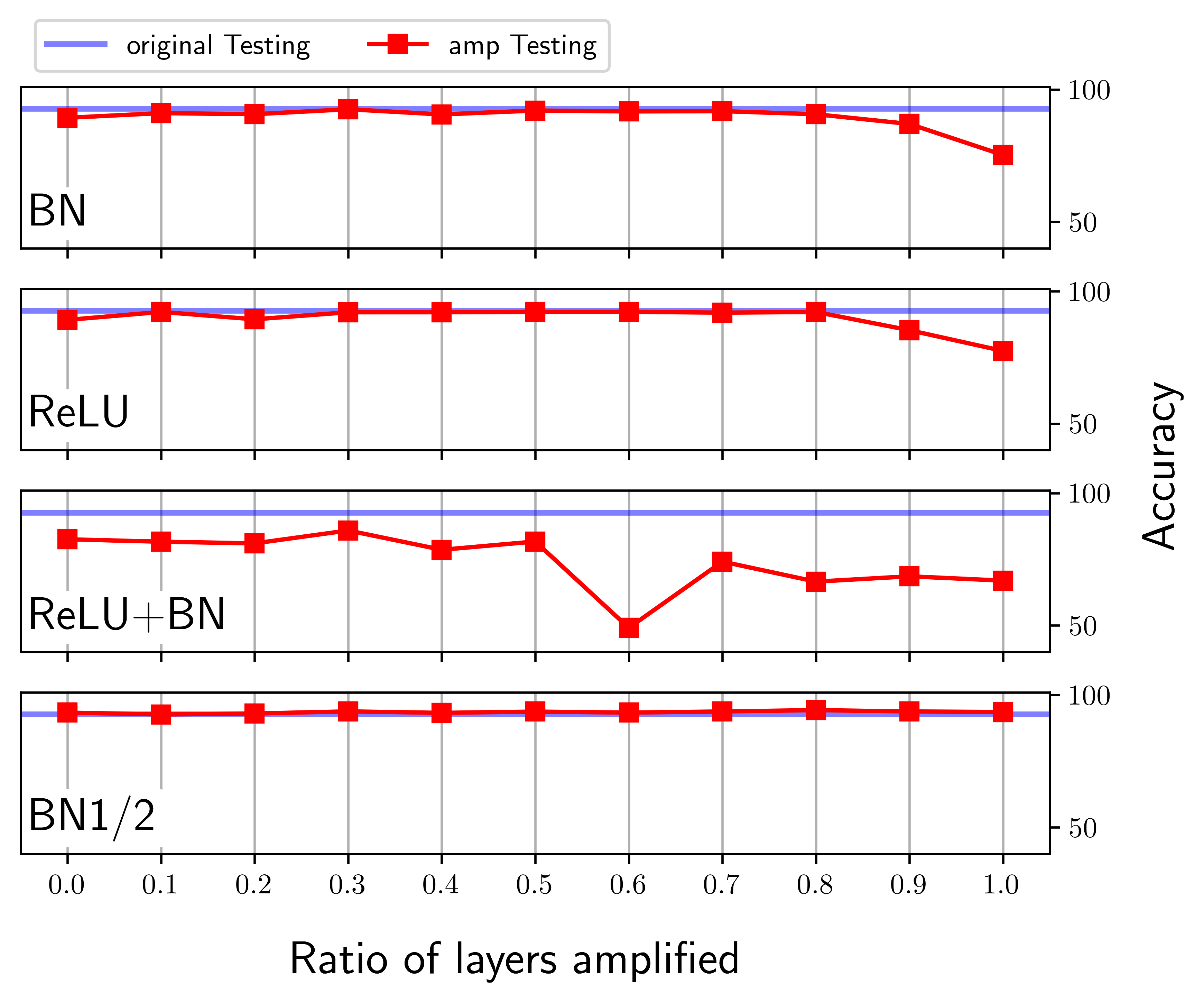}}
    \end{framed}
  \caption{Performance of the models after training with step-2 strategy with gradient amplification(red) applied from epochs 51-100 compared to mean accuracies of the original models(blue) with no gradient amplification. In each plot, blue horizontal line shows the average testing accuracy of the original models without gradient amplification. amp testing refers to testing accuracies of models with  gradient amplification. The type of the layer is shown in each subplot; horizontal and vertical axes correspond to the ratio of amplified layers and accuracies respectively. These experiment plots correspond to $params$  $S2\_0.5\_$\TTCLR{$xx$}, where the ratio(0.5) of layers are amplified for epochs 51-100. The other $params$ $S2\_0.1\_$\TTCLR{$xx$}, $S2\_0.3\_$\TTCLR{$xx$} and $S2\_0.6\_$\TTCLR{$xx$} also have similar performance patterns.}
  \label{fig:step2_all}
\end{figure*}

\subsubsection{Phase-1: (Effect of type of layers)} \label{subsec:effect_phase1}
In this work, ReLU, BN or both(ReLU+BN) are the layers used for gradient amplification. We run original models without gradient amplification 5 times and record their training, testing accuracies and compare the corresponding gradient amplified models with the mean of the these accuracies across 5 runs. For each type(s) of layer chosen, experiments are run for $params$ $S2\_0.1\_$\TTCLR{$xx$}, $S2\_0.3\_$\TTCLR{$xx$}, $S2\_0.5\_$\TTCLR{$xx$}, $S2\_0.6\_$\TTCLR{$xx$}, that is, for each ratio value in $\beta \in \{0.1, 0.3, 0.5, 0.6\}$ during epochs 51-100 ($\eta=0.1$), we build models by varying ratio values from $\{0.0, 0.1,..,1.0\}$ for epochs 101-130 ($\eta=0.01$), without amplification from 131-150 (see Fig. \ref{fig:GradAmp_Epochs_3}). The best training and testing accuracies of these models are compared with the average training and testing accuracies of corresponding original model. Since training accuracies of the original models are close to 100\%, we emphasize our comparison on testing accuracies. 

In VGG-19 model, we perform analysis considering ReLU, BN or both layers for gradient amplification and provide accuracy improvements for $params$ $S2\_0.1\_$\TTCLR{$xx$}, $S2\_0.3\_$\TTCLR{$xx$}, $S2\_0.5\_$\TTCLR{$xx$}, $S2\_0.6\_$\TTCLR{$xx$} respectively. When only ReLU layers are chosen, testing accuracies improve around $1.98\%$, $1.31\%$, $1.08\%$, $1.25\%$ respectively for above $params$. In the case of amplification applied only to BN layers, an improvement of $2.27\%$, $1.64\%$, $0.91\%$, $0.96\%$ in testing accuracies is observed.  When both ReLU and BN are chosen, models have accuracy difference of $0.9\%$, $0.64\%$, $0.13\%$, $0.3\%$ respectively. When both layers are used in amplification, the improvements across different models are less than $1\%$, which becomes better when only either ReLU or BN is used. Best improvements are seen when amplification is applied on only BN layers. 

Resnet models are made of residual blocks, each of which consists of two convolutional units and therefore each block has two ReLU and BN layers. In these models, other than experimenting with all ReLU and BN layers, we additionally perform experiments considering only one of the BN layers from residual blocks. When all the BN layers are considered for amplification in Resnet-18 models, there is an improvement of $1.66\%$, $1.35\%$, $0.53\%$, $0.34\%$  respectively for $params$ $S2\_0.1\_$\TTCLR{$xx$}, $S2\_0.3\_$\TTCLR{$xx$}, $S2\_0.5\_$\TTCLR{$xx$}, $S2\_0.6\_$\TTCLR{$xx$}. When only ReLU layers are used, there is a difference of $1.76\%$, $1.32\%$, $0.77\%$, $0.26\%$  respectively. When both BN and ReLU are used, there is an initial improvement of $01.14\%$, $0.03\%$, for $params$ $S2\_0.1\_$\TTCLR{$xx$}, $S2\_0.3\_$\TTCLR{$xx$} but the performance drops for $params$ , $S2\_0.5\_$\TTCLR{$xx$}, $S2\_0.6\_$\TTCLR{$xx$}. When only one of the BN layer from a residual block is considered, there is an improvement of $1.98\%$, $2.08\%$, $1.64\%$, $1.32\%$ for respective $params$. When both BN and ReLU are used for amplification, there is an improvement only for params $S2\_0.1\_$\TTCLR{$xx$} and either declines or slightly changes for the remaining $params$. When either ReLU or BN layers is considered, for params  $S2\_0.1\_$\TTCLR{$xx$}, $S2\_0.3\_$\TTCLR{$xx$}, more than performance improvement of more than $1.3\%$ can be observed and for params $S2\_0.5\_$\TTCLR{$xx$}, $S2\_0.6\_$\TTCLR{$xx$}, improvements are less than $1\%$. When one of the BN layers in residual blocks are considered, then the models have accuracy improvements of more than $1.3\%$ for all params and it also achieves the best testing accuracy of $94.57\%$ with an improvement of $2.08\%$ over original model. 

Similarly for Resnet-34, in the case of only BN layers, there is an accuracy gain of $1.21\%$, $0.64\%$ for $params$ $S2\_0.1\_$\TTCLR{$xx$}, $S2\_0.3\_$\TTCLR{$xx$} and then slightly decreases for $params$ $S2\_0.5\_$\TTCLR{$xx$}, $S2\_0.6\_$\TTCLR{$xx$} respectively. When only ReLU layers are considered, there is an improvement of $1.61\%$, $0.42\%$ for $S2\_0.1\_$\TTCLR{$xx$}, $S2\_0.3\_$\TTCLR{$xx$}, and for other $params$, there is a slight decrease(less than 0.8\%) for $params$ $S2\_0.5\_$\TTCLR{$xx$}, $S2\_0.6\_$\TTCLR{$xx$}. When both BN and ReLU are used, there is an initial improvement of $1.14\%$ for $params$ $S2\_0.1\_$\TTCLR{$xx$} and then it declines for other $params$.  When only one of the BN layers are used in a residual block, an improvement of $1.67\%$, $1.23\%$, $1.55\%$, $1.49\%$ can be seen for respective $params$. When all the BN layers are only used for amplification, there is an improvement of more than $1\%$ only for params $S2\_0.1\_$\TTCLR{$xx$} and the performance either declines or slightly changes for remaining $params$. Similar pattern is observed when only ReLU or both ReLU+BN are used for amplification. When one of the BN layers in residual blocks are considered, models have accuracy improvements of more than $1.2\%$ for all params and it also achieves the best testing accuracy of $94.39\%$ with an improvement of $1.67\%$ over original model. 

Our experiments show that for VGG-19 models, selecting ReLU improves the performance of the models, but achieves best performance when BN layers are chosen for amplification. In Resnet-18 and Resnet-34, performance of the models improve when BN layers are chosen for amplification and best performance is achieved when only of the BN layers from a residual block is selected for amplification.

\subsubsection{Phase-2: (Effect of ratio of selected layers $\beta$)}

Here, we discuss the impact of the ratio of selected layers on each of the above types. In our training strategy, gradient amplification is firstly applied in step-1 (as shown in Fig. \ref{fig:GradAmp_Epochs_2}) to determine the best performing ratio values for epochs 51-100. 
The best training and testing accuracies after gradient amplification across all the ratio values are compared with the original baseline models to analyze the overall effectiveness. In VGG-19, for all layer types, as the ratio of amplified layers increases, the performance of the model diminishes compared to original models. The training accuracies decrease at an increased rate compared to testing accuracies. When gradient amplification is performed, training and testing accuracies decrease slightly by $-0.3\% $ and $ -0.14\%$ (for BN only) and increase slightly by $0.02\% $ and $ 0.03\%$(in the case of ReLU+BN) and show an improvement of $0.65\% $ and $ 0.77\%$ (for ReLU only) respectively. 

In Resnet-18 and Resnet-34 models, as the ratio of amplified layers increases, training and testing accuracies remain close to the accuracies of the baseline models when only one of the BN layers in a residual block are considered. When either BN or ReLU is considered, as the ratio of amplified layers increases, performance of the models decreases slightly compared to original models. When both BN and ReLU are considered for amplification, as the ratio of layers increases, performance of the models decrease significantly compared to respective baseline models. In Resnet-18, the best training and testing accuracies after gradient amplification across all the ratio values, show an improvement by $0.69\% $ and $ 0.62\%$ (for BN only), $ 0.52\% $ and $ 0.67\%$ (for ReLU only), $0.52\% $ and $ 0.67\%$(in the case of ReLU+BN), and $0.79 \%$ and $ 0.92\%$ (in the case when one of BN layers from residual block) respectively. In Resnet-34, the best training and testing accuracies after gradient amplification across all the ratio values, show an improvement by $0.54\% $ and $ 0.68\%$ (for BN only), $ 0.4\% $ and $ 0.15\%$ (for ReLU only), $0.42\% $ and $ 0.81\%$(in the case of ReLU+BN), and $0.57 \%$ and $ 0.85\%$ (when one of the BN layers from residual blocks are considered) respectively.

In the case of step-1, amplification is applied only for epochs 51-100 ($\eta = 0.1$). We also perform experiments by applying amplification for epochs 101-150 ($\eta = 0.01$), by considering all or some of the epochs. We observe that the models perform better when amplification is applied from 51-100 epochs ($\eta = 0.1$) followed by 101-130 epochs ($\eta = 0.01$) as shown in step-2 (Fig. \ref{fig:GradAmp_Epochs_3}). To narrow the parameter space, we only consider ratio values for epochs 51-100 where the models perform better. From analysis of model performances in step-1, ratio values $\{0.1, 0.3, 0.5, 0.6\}$ on average provide better results and therefore, these ratios are used for epochs 51-100 ($\eta= 0.1$) as mentioned earlier and ratio values are varied for 101-130 epochs ($\eta= 0.01$), namely $S2\_0.1\_$\TTCLR{$xx$}, $S2\_0.3\_$\TTCLR{$xx$}, $S2\_0.5\_$\TTCLR{$xx$}, $S2\_0.6\_$\TTCLR{$xx$}. 
Fig. \ref{fig:step2_all} shows the performance of VGG-19, Resnet-18 and Resnet-34 models respectively for these params when different layers are amplified. Performance improvements of these models are discussed in Phase-1 in detail. Here we emphasize on the effect on the models as the ratio of amplified layers increases. For VGG-19 models, as the ratio of layers increases, there is an increase in performance initially  and then it decreases when the ratios above $0.7$. When both ReLU+BN are amplified, the models have significant decrease with the increase of ratio values. In the case of Resnet-18, models have improved or similar performance even as the ratio increases except when both ReLU+BN are amplified, in which case it decreases. For Resnet-34, models have improved or similar performance  even as the ratio increases until 0.8, after which it decreases. But when both ReLU+BN are amplified, the performance declines even for smaller ratios.

When amplification is applied using approach in step-1 (Fig. \ref{fig:GradAmp_Epochs_2}), the models perform better when only ReLU are amplified in the case of VGG-19 and for Resnet-18, Resnet-34, models perform best when only one of the BN layers from a residual block is used for amplification. When amplification is done as in step-2, all models achieve higher accuracies than baseline models for most of the ratio values except when ReLU+BN are amplified, in which case only some of the smaller ratio values have better models. This shows that a small ratio of amplified layers are sufficient to improve the performance of original models.

\subsubsection{Phase-3: (Effect of gradient amplification factor)}
\begin{figure}
\includegraphics[width=0.5\textwidth]{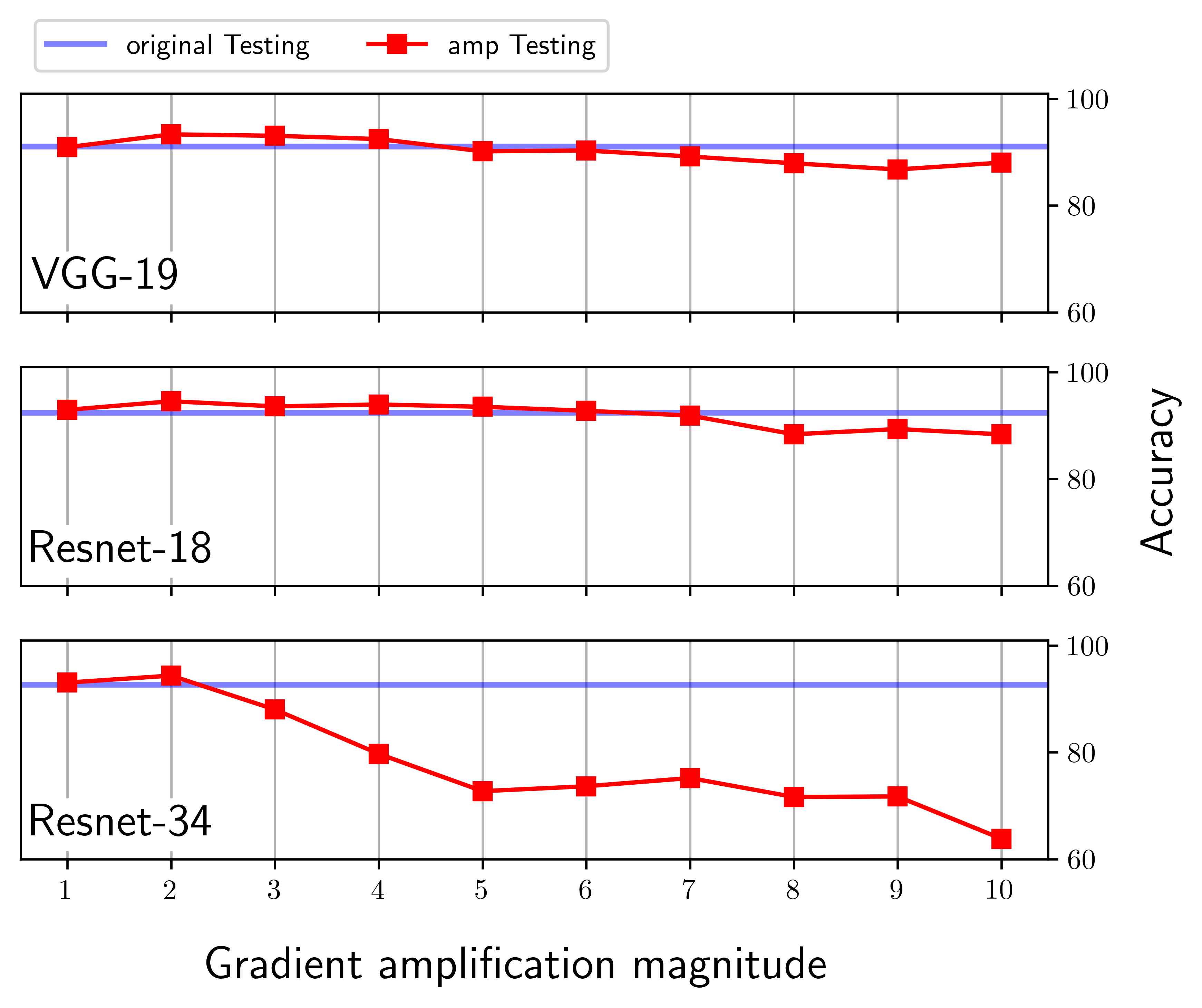}
\caption{ Performance comparison of amplified models(red) as $\Gamma$ is varied from 1 to 10 (horizontal axis) (vs) original models (blue).}
\label{fig:grad_model_results}
\end{figure}

Fig. \ref{fig:grad_model_results}  shows the performance of models as $\Gamma$ is varied. $params$ of the best models after analysis phase-1 and phase-2 are taken and gradients are amplified by varying the value of $\Gamma$ from $\{1,2,3,..,10\}$. For VGG-19, the best model is achieved while amplifying only BN layers for $params$ $S2\_0.1\_0.3$. For Resnet-18 and Resnet-34, the best models are achieved while amplifying only one of the BN layers in residual units and for $params$ $S2\_0.3\_0.5$ and $S2\_0.1\_0.7$ respectively. While changing values of $\Gamma$, as the factor of amplification $\Gamma$ increases, the performance of the models declines. To generalize, we can say that when $\Gamma$ is more than 5, the models do not perform better or sometimes perform worse than the corresponding baseline models. Effect of amplification factor $\Gamma$ also depends on the ratio of layers being amplified. If the ratio is close to 1, then $\Gamma$ values less than 5 can also decrease performance of the models. 

\begin{figure}
\includegraphics[width=0.5\textwidth]{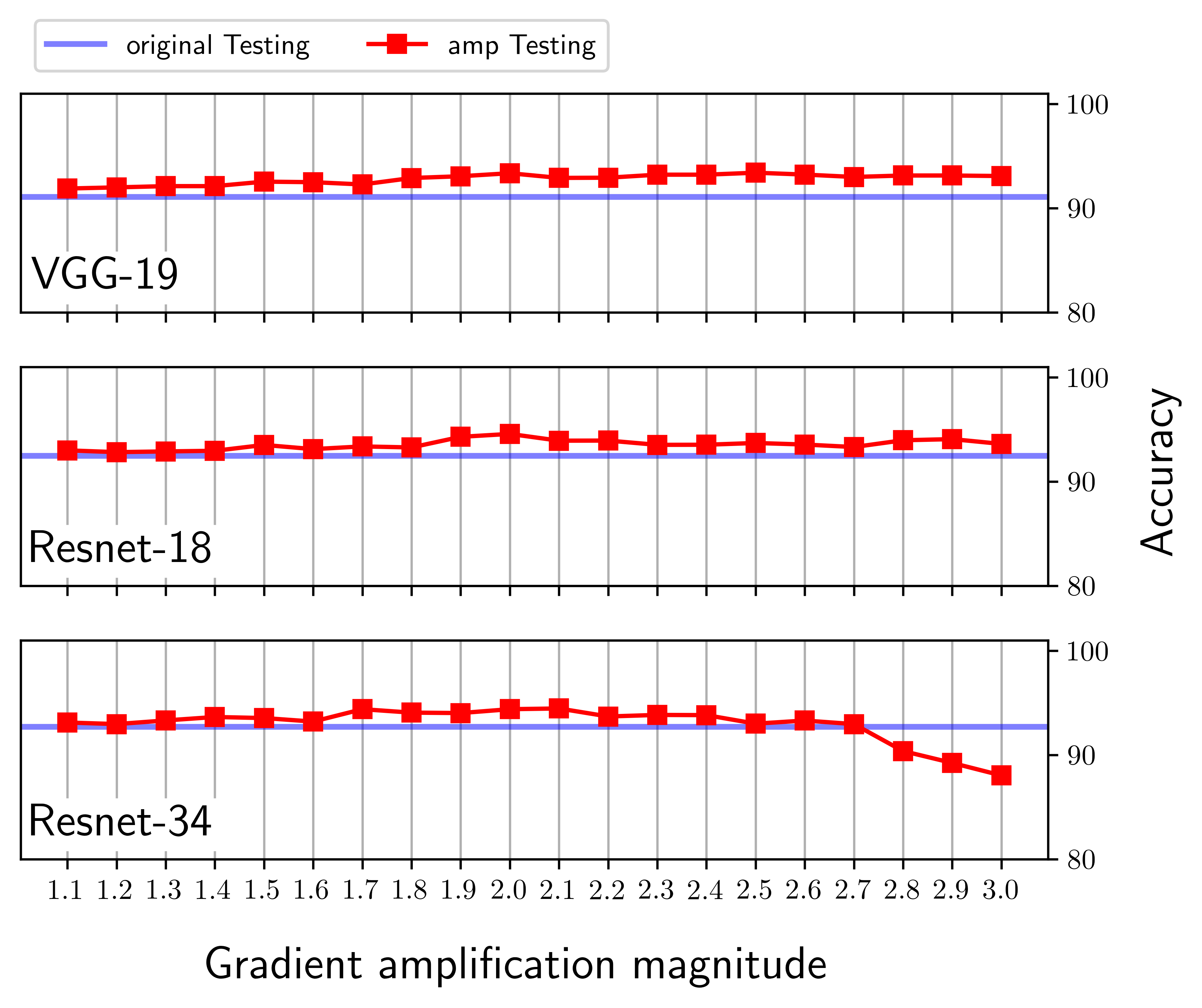}
\caption{ Performance comparison of amplified models(red) as $\Gamma$ is varied in small steps from 1 to 3 (horizontal axis) (vs) original models(blue).}
\label{fig:grad_step_model_results}
\end{figure}

We also perform experiments by fine-tuning the amplification factor from 1 to 3 in steps of 0.1, i.e., by varying $\Gamma \in \{1.1,1.2.1.3,1.4,....,3\}$. Fig. \ref{fig:grad_step_model_results} shows the performance of these models as $\Gamma$ is varied in small steps from 1 to 3. In the case of VGG-19 and Resnet-18, the model always performs better than the baseline models both during training and testing and for Resnet-34, the model performs better until 2.7 and declines after that. In all these models, it can be observed that the best accuracy is around the value 2 which justifies our experiment analysis in the above phases.

\begin{figure*}
    \begin{framed}
  \centering
    \subfigure[VGG-19\label{fig:epoch-vgg-19}]{\includegraphics[width=0.33\textwidth]{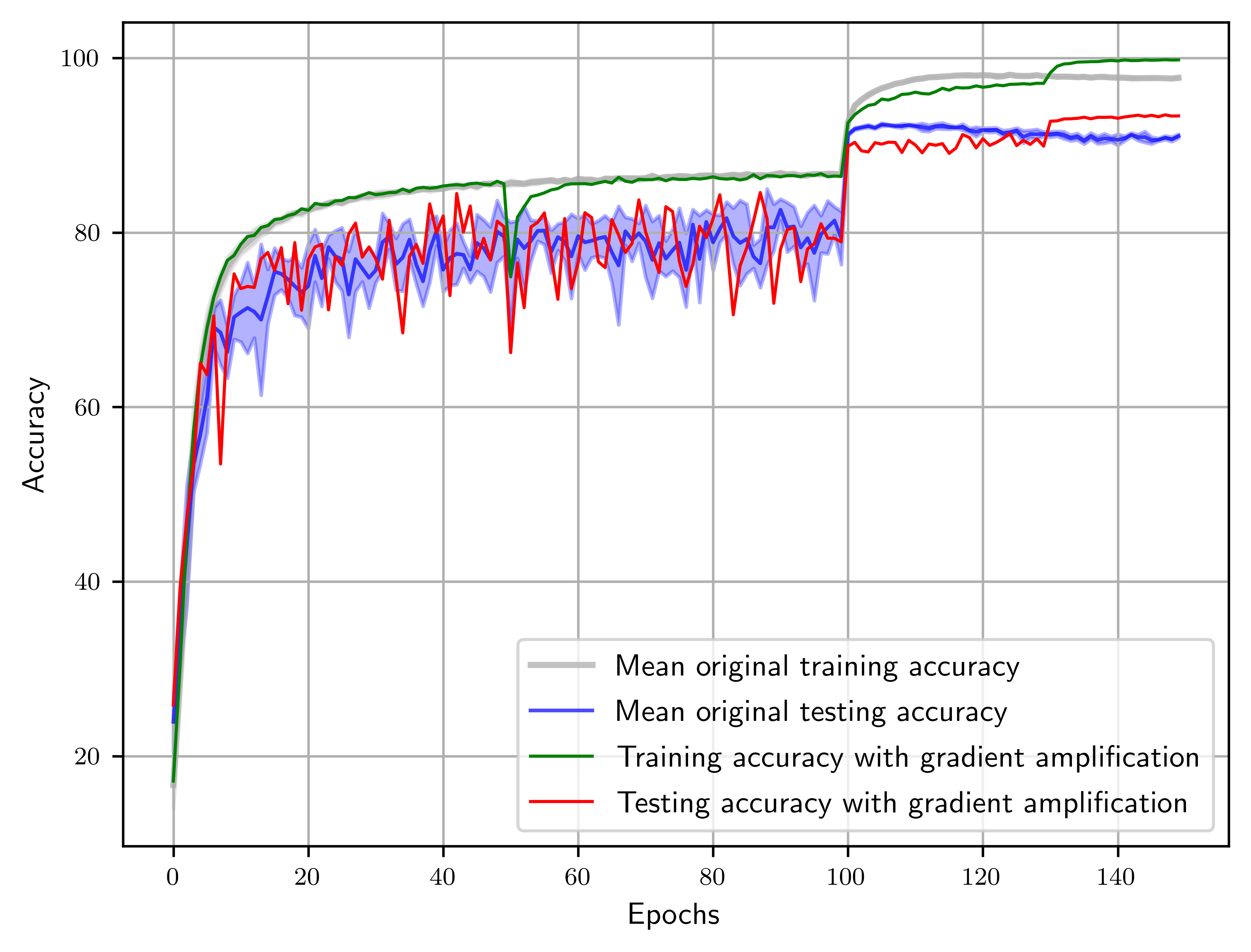}}
    \subfigure[Resnet-18\label{fig:epoch-resnet-18}]{\includegraphics[width=0.33\textwidth]{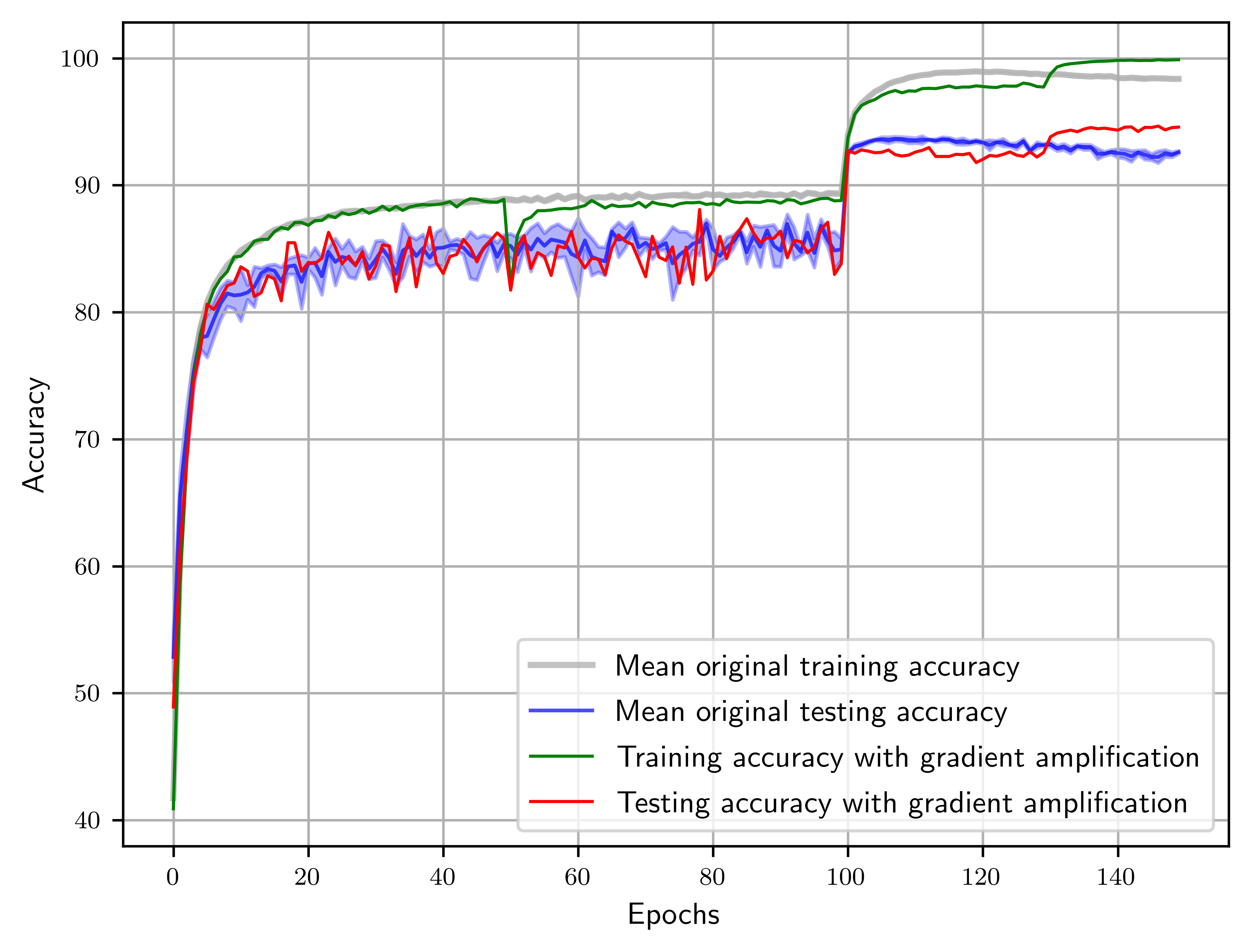}}
    \subfigure[Resnet-34\label{fig:epoch-resnet-34}]{\includegraphics[width=0.33\textwidth]{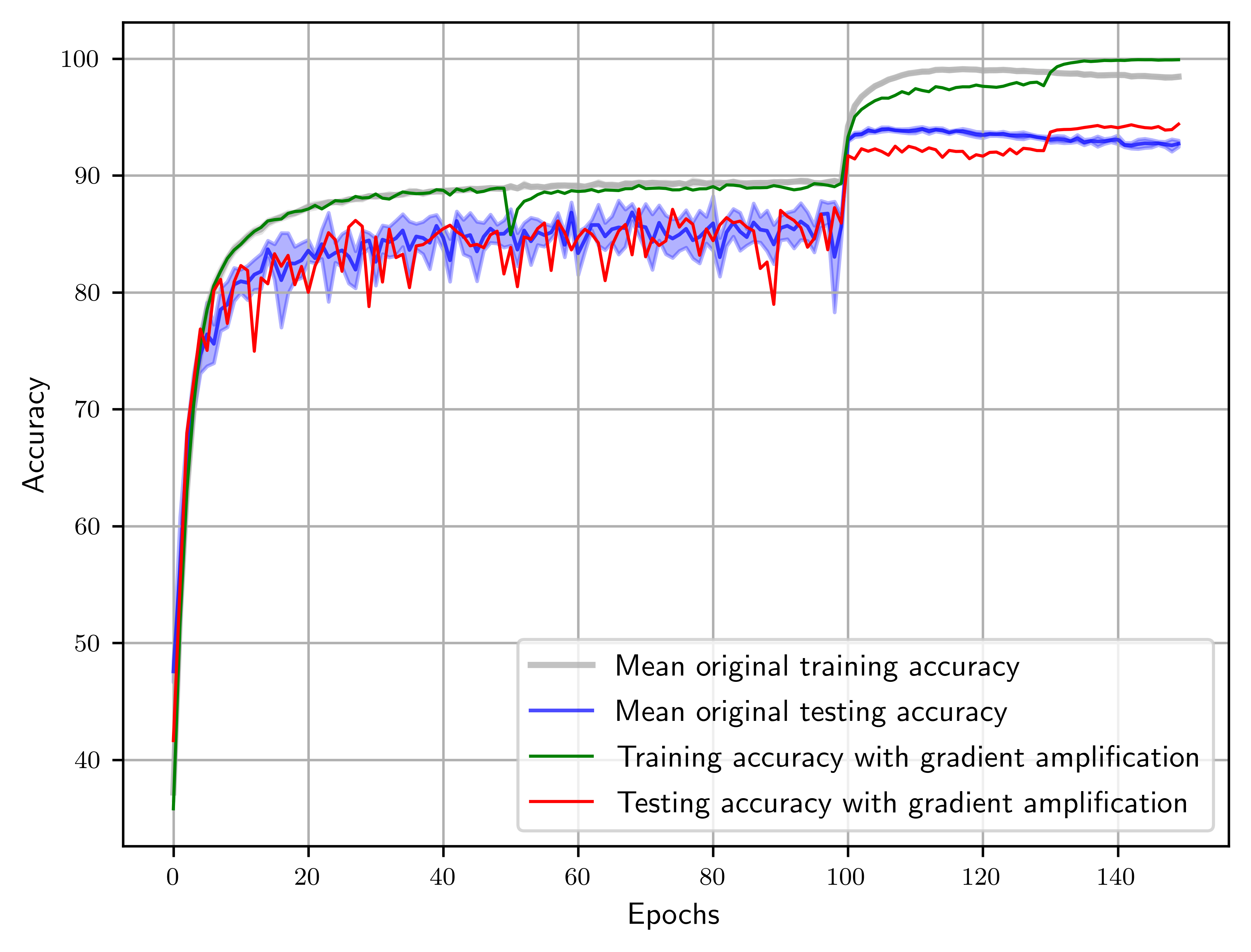}}
    \end{framed}
  \caption{Performance of the best models with gradient amplification over 150 epochs compared to original model with no gradient amplification. Original training(gray), testing(blue) accuracies including their mean accuracies are plotted along with amplified training(green) and testing(red) accuracies. These plots demonstrate that the models do not overfit while training with amplification.
 }
  \label{fig:best_model_epochs}
\end{figure*} 

\subsection{Best models}

The best performance of all the models is shown in Table \ref{tab:best_models}. Performance improvements can be observed both in  
 and testing accuracies. The rows with `original' in $params$ column show the performance of the original model with no gradient amplification. The following grayed row shows the performance of the corresponding model with gradient amplification. The $params$ that achieve these best models are also shown. We can observe that gradient amplification increases both training and testing accuracies. Though training accuracies are very close to 100\% in the original model, gradient amplification improves them further. It can be noted that resnet models comprises of residual blocks architecture (an extra connection to the preceding layer) which already overcomes vanishing gradients by passing the current gradients directly to the previous layers without modification using residual connection, and therefore an improvement of $1.67\%$ can be assumed to be significant. 

\definecolor{LightCyan}{rgb}{0.88,1,1}
\definecolor{Gray}{gray}{0.85}
\def\mathbi#1{\textbf{\em #1}}

\newcommand{\CC}[1]{\cellcolor{Gray}}

\begin{table}[h]
  \caption{Accuracy comparison of models with gradient amplification (vs) mean accuracies of corresponding original model across 5 runs  }
  \label{tab:best_models}
  \centering
  \resizebox{.48\textwidth}{!}{%

\renewcommand{\arraystretch}{1.4}
  \begin{tabular}{|c|l|l|l|l|l|}
    \hline
   \multirow{2}{*}{Model} &\multirow{2}{*}{$params$}&\multicolumn{2}{c|}{Mean/Best accuracy (\%)}&\multicolumn{2}{c|}{Improved accuracy (\%)}\\
   \cline{3-6}
            & &Training       &Testing &Training       &Testing\\
    \hline
    & Original & 97.87 & 91.08 & -- & -- \\
    \multirow{-2}{*}{VGG\_19}
    &\CC{50}\begin{minipage}[t]{0.21\columnwidth}%
$Ours$ ($VGG\_19$ with amplification)
\end{minipage}  &\CC{50}99.764 &\CC{50} 93.35 &\CC{50}1.9 & \CC{50}\textbf{2.27} \\
\hline
    & Original & 98.371 & 92.488 & -- & -- \\
    \multirow{-2}{*}{Resnet\_18}
    & \CC{50}\begin{minipage}[t]{0.21\columnwidth}%
$Ours$ ($Resnet\_18$ with amplification)
\end{minipage}  & \CC{50}99.878 & \CC{50}94.57 & \CC{50}1.51 & \CC{50}\textbf{2.08} \\
    \hline
    
    & Original & 98.444 & 92.716 & -- & --\\
    \multirow{-2}{*}{Resnet\_34}
    & \CC{50} \begin{minipage}[t]{0.21\columnwidth}%

$Ours$ ($Resnet\_34$ with amplification)
\end{minipage}  & \CC{50} 99.774 & {\CC{50}}94.39 &{\CC{50}}1.25 & {\CC{50}}\textbf{1.67} \\

    \hline
  \end{tabular}
  }
\end {table}

Fig. \ref{fig:best_model_epochs} shows the performance of the each of the models for the params listed in Table \ref{tab:best_models}. These plots demonstrate that the models trained with gradient amplification do not cause overfitting problem. In the case of VGG-19, the best model is achieved when amplification is applied only on BN layers and for Resnet models, the best models are achieved when only one of the BN layers from a residual block are considered for amplification. Gradient amplification model surpasses the performance of all the original models. Accuracies achieved by amplified models not only exceed the mean average accuracies across 5 runs of the original models, but also outperform the best accuracy among these 5 runs of the original models. 

To analyze the impact of training time, we perform experiments on the original models without amplification. Even the best models out of 5 runs do not achieve performance close to the corresponding amplified models. We ran all the original models for 50 more epochs with the learning rate of 0.01(i.e., 151-200 with $\eta=0.01$) and notice that these models do not reach the performance of amplified models. We need to reduce the learning rate further for the next epochs to improve performance of the original models and therefore there is no direct way of comparing them. But clearly, our proposed method of training with amplified gradients can train the deep learning models at higher learning rates to achieve better performance. Original models take more time to achieve similar accuracy (at the same settings) as amplified models.


\section{Conclusions \& Future Work}\label{sec:conclusion}
In this work, we propose a novel gradient amplification method to dynamically increase gradients during backpropagation. We also provide a training strategy consisting of set of epochs with switching between gradient amplification and without amplification. Detailed experiments are performed on VGG-19, Resnet-18 and Resnet-34 models to analyze the impact of gradient amplification with different amplification parameters. We learn that only a proportion of layers are sufficient to attain such a performance gain. It can also be observed that BN layers give the best improvement while performing amplification, followed by ReLU layers, whereas the performance quickly diminishes when both ReLU+BN are used. All these experiments show that our proposed amplification method and training strategy increase the performance of the original models and achieve better accuracies even at higher learning rates. In future work, we would like to perform the experiments on the larger datasets having more output classification classes like CIFAR-100 and ImageNet. In our current experimental models, there were no exploding gradients. However, we would also like to explore other models having such an issue and experiment gradient diminishing dynamically, similar to graident amplification, to address gradient exploding problem.


%



\ifCLASSOPTIONcompsoc
  \section*{Acknowledgments}
\else
  \section*{Acknowledgment}
\fi

The authors would like to thank Georgia State University(GSU) for providing us access to  High Performance Computing(HPC) cluster on which most of the experiments are performed. This research is supported in part by a NVIDIA Academic Hardware Grant.

\ifCLASSOPTIONcaptionsoff
  \newpage
\fi

\balance
\bibliography{project}
\bibliographystyle{IEEEtran}

\end{document}